# Offence Detection in Dravidian Languages using Code-Mixing Index based Focal Loss

Debapriya Tula*, Shreyas MS, Viswanatha Reddy, Pranjal Sahu, Sumanth Doddapaneni, Prathyush Potluri, Rohan Sukumaran, Parth Patwa*

IIIT Sri City, India
*Corresponding authors
E-mail: debapriya.t17@iiits.in, parthprasad.p17@iiits.in

**Abstract** Over the past decade, we have seen exponential growth in online content fueled by social media platforms. Data generation of this scale comes with the caveat of insurmountable offensive content in it. The complexity of identifying offensive content is exacerbated by the usage of multiple modalities (image, language, etc.), code-mixed language and more. Moreover, even after careful sampling and annotation of offensive content, there will always exist a significant class imbalance between offensive and non-offensive content. In this paper, we introduce a novel Code-Mixing Index (CMI) based focal loss which circumvents two challenges (1) code-mixing in languages (2) class imbalance problem for Dravidian language offense detection. We also replace the conventional dot product-based classifier with the cosine-based classifier which results in a boost in performance. Further, we use multilingual models that help transfer characteristics learnt across languages to work effectively with low resourced languages. It is also important to note that our model handles instances of mixed script (say usage of *Latin* and *Dravidian - Tamil* script) as well. To summarize, our model can handle offensive language detection in a low-resource, class imbalanced, multilingual and code-mixed setting.

**Keywords** Multilingual · Offense Detection · Deep Learning · Code-Mixing · Loss Function

## 1. Introduction

Communication has never been as sophisticated as it is today. Online communication has helped break a ton of obstructions as far as time, distance and simplicity of correspondence are concerned. As the active users of the internet filled in volumes throughout past few years prompted an extraordinary expansion in the measure of hate speech in the open web. Online hate speech has been connected to a rise in violence against minorities around the world, including mass shootings, lynching, and ethnic genocide [23]. As more people go online, people who are prone to exhibit racism, misogyny, or homophobia have found niches that might reinforce their beliefs and provoke violence. Hate speech is frequently pointed towards a group or an individual, harming their character, convictions, identity and/or feelings. Many rogue elements use these platforms to revile and cause disharmony locally as offensive content, at many times, are inseparable from the benign posts. The absence of rigorous checking adds to the opportunity. It is thus important to solve this issue. Social media is easily accessible to a large domain of people and its scale restricts us from manually monitoring and filtering the content shared. Given the volume of data on social media, artificial intelligence (AI) must be a part of the regulatory mix and this calls for the need for (semi) automatic systems for the identification of hate or offensive speech.

The style of data on social media also plays a major role in understanding the data. The language structure is regularly absent and individuals will generally utilize words from various languages, ultimately resulting in code-switched data [3, 32]. The issue is exacerbated as people use words from various contents, say blending both Latin and native scripts (*Devanagari, Dravidian, Mandarin etc*) in their language. A unified model which can understand a multitude of these scripts plays a significant role in understanding the discourse in social media data and is conducive to creating a safer and healthier digital space.

The significance of the issue and the difficulties presented call for novel ideas for offensive language detection. Owing to this, multiple workshops [1, 51] and shared tasks [7, 20, 22] have been conducted to address the problem at hand.

In this paper, we extend the previous work [48], which presented a deep learning-based system for offensive language identification in Dravidian languages. Here, we introduce a novel Code-Mixing Index (CMI) based focal loss to handle the code-mixing. We further use it with Cosine Normalization to allow the classifiers to have more balanced decision boundaries i.e. more space for classes with less number of examples.

## 2. Related Works

With an increase in ease of access to social media, there is also an increase in evidence of usage of code-mixed language online. Authors in [2] found that almost 17% of the comments in a dataset of Facebook posts were multilingual. Initial attempts to analyze the sentiment in the context of Indian language tweets were made by [31] where the authors focused on the classification of the polarity of tweets in Bengali, Hindi, and Tamil languages. Vyas et. al. [50] attempted pos-tagging of code-mixed social media posts. Other studies which address various aspects related to code-mixing like language identification, language modelling, word-level language identification etc. include [3, 9, 11, 35].

Offensive language detection has been an important area of research with the growing number of internet users. This has been studied in varied formats like aggression detection, cyberbullying, offensive language etc. Fortuna and Nunes, 2018 [14] discuss the nuances of hate speech and its potential for negative cultural effects, specifically in online communities and social media platforms. Wherein [46] investigated abusive language detection in a multilingual setting for English and German.

Research on offensive language detection is more important than ever, as demonstrated by numerous shared tasks and research works recently [4, 7, 20, 21, 22, 33].The SemEval 2019 Task 5 [4] focused on Hate Speech detection on Immigrants and Women on Twitter. They organized two subtasks i.e. a binary task for identification of hate speech and a fine-grained task for further identification of target group and aggression attitude. They created a new dataset with 19.6k tweets spanning two languages English and Spanish. Task 6 [52] focused on the identification and categorization of offensive language in social media. The task had a similar sub-division of tasks i.e. a binary task for classification, offensive type classification and target of the offensive content. They proposed a new dataset, OLID, which has 14k tweets annotated using a hierarchical annotation model. OffensEval 2020 [53] was a profanity identification task presented in SemEval 2020. It was conducted in 5 languages - English, Arabic, Danish, Greek, and Turkish.

There are various works on offensive language and hate speech detection using machine learning and deep learning algorithms i.e. CNN, RNN, LSTM, *etc.* [29, 45]. With the advent of Transformer [41] and BERT based models [12, 18, 44] there has been an increase in their usage in this field.

The authors in [45] used an ensemble of CNN and RNN based models for offensive language detection in German Tweets. The authors in [39] accomplished the task of Robust Aggression Identification describing an ensemble of multiple fine-tuned BERT models based on bootstrap aggregating (Bagging). Transfer learning capabilities can be augmented using Task Adaptive pre-training on BERT models [37]. To alleviate the data imbalance and low-resource issues, [25] proposed a generative based augmentation technique.

Further, word-level recognition of code-mixed data is used in emotion classification for sentiment analysis [9]. [27] proposed an approach based on a weighted loss for multilingual models focusing on the complexity of code-mixing sentences, concluding that the eccentricity of word representations used has a considerable impact on the performance of a model.

There have been advances in offensive content detection in German [40], Italian [10], English-Hindi [30, 34], Bengali [42] but there has been little progress has beenmade in Dravidian languages. Recent efforts include LSTM [28], Transformer [13] based methods for offensive content detection in Dravidian languages. In [38], authors experiment with various inter-task and multi-task transfer learning techniques to leverage the useful resources available for offensive speech identification in the English language and enhance the multilingual models with knowledge transfer from related tasks.

## 3. Data

Three languages are considered for our work: Kannada [17], Malayalam [5], and Tamil [6]. The six class labels in Kannada and Tamil are:
– Not Offensive - (NO)
– Not Native - (NN)
– Offensive Individual - (OI)
– Offensive Group - (OG)
– Offensive Untargeted - (OU)
– Offensive Other - (OO)

All the above-mentioned classes except 'Offensive Other' are present in the Malayalam dataset. The distribution of the data is described in Table 1 for all three languages. In total there are 5936, 11695, 34898 samples for Kannada, Malayalam and Tamil respectively. The values of Krippendorff's alpha, which signifies the inter-annotator agreement, for Kannada, Tamil and Malayalam are 0.78, 0.66, 0.89 respectively.

There was a significant class imbalance in all 3 languages in the dataset. 'Not Offensive' class, which is the majority class in all three languages, accounts for 56.97% of the samples in Kannada, 88.77% of the samples in Malayalam and 72.25% of the samples in Tamil data. In order to overcome this skewness, we utilize class weighting to penalize more for the under-represented classes. This is discussed in detail in the next section.

We often see the same native word written in different ways in English. The word "your" in Kanada is writen as ನಿನ್/ನಿನ್ನ ನಿಮ್/ನಿಮ್ಮ nin/ninna (singular) and nim/nimma (plural). nin/ninna (singular) and nim/nimma (plural).

| Class | Kannada | Malayalam | Tamil |
|---|---|---|---|
| NO | 3382 | 10382 | 25215 |
| NN | 1407 | 882 | 1447 |
| OI | 486 | 171 | 2338 |
| OG | 327 | 106 | 2550 |
| OU | 212 | 154 | 2894 |
| OO | 122 | - | 454 |
| Total | 5936 | 11695 | 34898 |

**Table 1** Data distribution of the three datasets.

On sampling and manually validating a few examples from the dataset, we noticed some instances in the Kannada data which have un- likely ground truth or could be perceived differently than how the annotators did. However, we did not make any changes to the labels. Some examples:

- *Krishana shapa tatteleebeku* (Krishana should curse (you)) is marked not offensive, however, the intent of this is to offend someone.
- *Nanu yash sudeep darshan appu cinema bittu yardu nodalla* (I only watch Yash, Sudeep, Darshan and Sudeep's movies) seems not offensive but is marked "Offensive_Targeted_Insult_Other".
- Nija Film ast channagidya nanu nodide nange kandita arta agle Illa (Is the film really that good? I saw it and honestly didn't understand it), marked Not-offensive, but could be perceived as offensive.

## 4. Methodology

This section outlines our approach to detect (classify) offensive content in Dravidian languages. Given the presence of both Latin script and text in the native language, we tackle this problem from the multi-lingual learning space. We employ large language models that are pre-trained on multiple languages, including Dravidian languages. Further, because of the extreme class-imbalance present in the data, we use the Focal loss along with class weighting to mitigate the model biases towards a particular dominant class. Owing to the code-mixed data presence, we propose a novel loss function - CMI-FL which integrates Code-Mixing Index (CMI) with focal loss. Further, we enhance our results by utilizing cosine normalization and pseudo-labelling.

### 4.1 Model

We leverage the multilingual transformer-based models DistilmBERT [44] and IndicBERT [18].

#### 4.1.1 DistilmBERT

DistilmBERT [44] has the same general architecture as BERT [12], however, is only about 60% the size of BERT as the number of layers is reduced by a factor of 2. The model is trained with the same pre-training objective as BERT i.e masked language modelling and next sentence prediction. The DistilBERT is 60 % fasterthan the BERT and retains almost 97 % of BERT's language understanding capabilities.

The efficacy of the DistilBERT inspired us to use a distilled version of the BERT base multilingual model (mBERT-base) called the *DistilBERT-base-multilingual-model*. DistilmBERT is about twice as fast as mBERT-base, according to HuggingFace[1]. The model has 6 layers, 768 dimensions and 12 heads, totalizing 134M parameters compared to 177M parameters for mBERT- base. For our work, we use the *cased model* as the data is code-mixed with English (the only case sensitive language in the corpora). The model was pre-trained on the concatenation of Wikipedia data in 104 different languages including Tamil, Malayalam and Kannada.

#### 4.1.2 IndicBERT

IndicBERT [18] is an ALBERT based model trained exclusively on Indian languages. The model is pre-trained on 11 Indian languages (including Dravidian languages- Kannada, Telugu, Malayalam, Tamil), and English using the standard Masked Language Modelling (MLM) objective. The model is pre-trained on news articles, magazines and blog posts.

### 4.2 CMI-FL: Code-Mixing Index based Focal Loss

The data is code-mixed, which adds to the difficulty of the classification task. To incorporate the difficulty due to code-mixing, we propose a novel loss function called Code-Mixing Index (CMI) loss function. It is a modification of focal loss [24], like [27]. Here, the focal loss is weighed by the level of code-mixing in a sentence, which is given by the Code-Mixing Index (CMI) [15]. The loss function $L$ is calculated as:

$$L = \alpha * CE * (1 - CMI)^{\gamma} + \alpha * CE * CMI^{\gamma}$$

where CMI is the Code-Mixing Index, CE is the initial cross-entropy, $\alpha$ and $\gamma$ are the constant positive scaling factor and the focusing parameter respectively. The word-level annotations for the datasets required by the CMI are approximated by using the pyenchant[2] tool, which provides language identification, spellchecking and a host of other capabilities. This CMI loss function can be used on tasks beyond the Dravidian languages.

---

[1] https://huggingface.co/distilbert-base-multilingual-cased#model-card

[2] https://pyenchant.github.io/pyenchant/

### 4.3 Class Weighting

As mentioned in section 3, the class imbalance is observed in the dataset, which can skew the model's predictions against under-represented classes. We use an inverse weighting strategy which helps penalize the under-represented classes more in the loss function.

### 4.4 Pseudo-Labelling

Pseudo-labeling is a semi-supervised learning technique where instead of manually labelling the unlabeled data, approximate labels are given on the basis of labelled data using a pretrained model. The model is first trained over the small set of labelled examples, which is then used to predict the labels for test data and this pseudo-labelled (test) data is used along with the train set for training the model further. This results in a considerable boost in performance.

### 4.5 Cosine Normalization

Cosine classifier has demonstrated encouraging results in vision-based applications like few/zero-shot learning and long-tail recognition. Inspired by these results we adopted a cosine similarity-based classifier [16, 26, 36] to obtain the final classification scores. We use cosine similarity-based classifier instead of dot product based classifier when computing the class activation scores. We normalize the final input embedding ($e$) i.e., the embedding just before the classification layer, as well as the classifier's weight vectors $\{wj\}_{j=1}^{C}$. Here $C$ corresponds to the number of classes, Bias is not used for the classifier.

$$e_{norm} = \frac{||e||^2}{1 + ||e||^2}$$

$$W_{norm} = \frac{w_C}{||w_C||^2}$$

Here $W_C$ corresponds to classifier's weight matrix. The matrix multiplication is performed on the two matrices normalized input embedding ($e_{norm}$) and normalized weight matrix ($W_{norm}$) to obtain the final classification logits. The input embedding's normalization is a non-linear squashing function adopted from [26, 41], which ensures that small-magnitude vectors are reduced to almost zeros, while large-magnitude vectors are normalized to a value slightly below one. Balancing the norms leads to more balanced decision boundaries, al-lowing the classifiers for few-shot classes to occupy more space [19]. In this work, we check the effectiveness of the cosine-based classifier using multiple loss functions, such as cross-entropy loss, CMI loss, and focal loss. The results are tabulated in table 2.

## 5. Experiments

Here, we describe the experiments performed. The models are trained on colab[3]. The code will be made public.

### 5.1 DistilmBERT

The text is tokenized using the DistilmBERT tokenizer having a vocabulary size of 110k. The maximum sequence length used is 128. The long input text is truncated and shorter sequences are handled by padding special tokens. A batch size of 8 is used for training the model for 10 epochs. The model's weights are optimized by using the Adam optimizer with a learning rate of 1e-8.

### 5.2 IndicBERT

For pre-processing, the IndicBERT tokenizer having a vocabulary size of 200k is used. A max sequence length of 200 is used and input sequences are truncated or padded depending on their length as in DistilmBERT. For our downstream task of offence detection, we add a fully connected layer with a dropout of 0.3 on top of the AlBERT model. The model is trained using the pre-trained IndicBERT weights only. A batch size of 32 is used during training, while a batch size of 16 is used for validation. The model was optimized using the Adam optimizer with a learning rate of 1e-5.

### 5.3 Code-Mixing Index (CMI) loss

For labelling the languages at the word level, we use the polyglot [8] module, known to support many multilingual applications. We calculate the amount of code-mixing of native language with English and assign a CMI for the entire batch of data at a time.
$α$=1.7 (constant positive scaling factor) and $γ$=0.25 (focusing parameter) are fixed for all experiments.

## 6. Results and Analysis

Table 2 shows the results of all the experiments over all the languages. We can see from the comparisons between using Cross-Entropy (CE) loss, CMI loss and Focal loss that the proposed CMI loss provides the best or comparable-to-best results due to its nature as described in Section 4.5. We implement Cosine Normalization and compare its performance when it is used along with Cross-Entropy loss, CMI loss and Focal Loss. It improves the results in most cases and gives a considerable boost to CMI loss for Kannada and Malayalam languages. We compare CMI loss + Cosine Normalization with and without pseudo-labelling and see that pseudo-labelling improves the performance. Please refer to [40] for more results of pseudo-labelling and class weighting. CMI loss along with Cosine Normalization and Pseudo Labeling achieves the best results for the Malayalam and Kannada languages in terms of all the metrics and is close to the best results for Tamil.

---
[3] https://colab.research.google.com/

|  | Kannada | | | Malayalam | | | Tamil | | |
| --- | --- | --- | --- | --- | --- | --- | --- | --- | --- |
|  | Precision | Recall | F1 | Precision | Recall | F1 | Precision | Recall | F1 |
| Distilmbert + CE loss + PL | 66.5 | 69.1 | 67.5 | 96.8 | 96.8 | 96.8 | 74.6 | 77.0 | 75.2 |
| Distilmbert + Focal loss + PL | 65.8 | 68.4 | 66.9 | 96.1 | 96.2 | 96.1 | **75.1** | 76.8 | **75.8** |
| Distilmbert +CMI loss + PL | 67.5 | 69.1 | 67.9 | 96.4 | 96.4 | 96.3 | 74.6 | **77.3** | 75.6 |
| Distilmbert +CE loss + CN + PL | 65.9 | 68.2 | 66.7 | 96.2 | 96.3 | 96.2 | 73.7 | 76.4 | 74.8 |
| Distilmbert +Focal loss + CN + PL | 66.5 | 68.5 | 67.3 | 96.6 | 96.8 | 96.7 | 73.7 | 73.4 | 74.8 |
| Distilmbert +CMI loss + CN | 67.0 | 68.6 | 67.4 | 96.4 | 96.5 | 96.3 | 74.9 | 76.3 | 75.5 |
| **Distilmbert + CMI loss + CN +PL** | **68.5** | **70.3** | **69.1** | **97.0** | **97.1** | **96.9** | 74.4 | 77.1 | 75.6 |
| IndicBert + CE loss | 66.5 | 65.1 | 65.6 | 93.9 | 93.9 | 93.9 | 60.8 | 65.4 | 62.3 |
| IndicBert + CE loss + CN | 63.5 | 60.4 | 60.6 | 95.9 | 95.7 | 95.8 | 68.9 | 69.7 | 69.2 |
| IndicBert + CMI loss + CN | 67.0 | 64.7 | 65.6 | 95.4 | 95.5 | 95.4 | 69.3 | 70.0 | 69.3 |
| IndicBert + CMI loss | 61.3 | 59.6 | 60.2 | 93.7 | 94.4 | 93.9 | 70.0 | 69.9 | 69.8 |

**Table 2** Results of all the experiments over all languages. All values mentioned are in percentage. All experiments use Class Weighting. CE = Cross entropy, CN = Cosine Normalization. PL = Pseudo labeling.

|  | NO | OTIG | OTII | OTIO | OU | NK |
| --- | --- | --- | --- | --- | --- | --- |
| Support | 426 | 45 | 66 | 16 | 33 | 191 |
| CE loss + PL | 76.8 | 28.9 | 52.4 | 8.7 | **17.02** | 74.5 |
| Focal loss +PL | 77.1 | 26.7 | 56.5 | 9.5 | 0.0 | 73.9 |
| CMI loss + PL | 77.8 | **34.3** | 55.2 | 20.0 | 8.0 | 72.9 |
| CE loss + CN + PL | 77.0 | 21.7 | 53.3 | 10.0 | 15.3 | 72.6 |
| Focal loss + CN + PL | 77.9 | 32.0 | 53.1 | 9.0 | 9.8 | 71.4 |
| CMI loss + CN | 77.1 | 29.9 | 55.9 | 14.8 | 6.9 | 73.7 |
| **CMI loss + CN + PL** | **78.4** | 32.5 | **58.0** | **28.5** | 12.2 | 73.9 |

**Table 3** Class-wise F1 scores of experiments with the distilmbert model for the Kannada dataset. All values reported are percentage. All experiments use Class Weighting. CE = Cross entropy, CN = Cosine Normalization. PL = Pseudo Labeling. Cl NO = Not_offensive, OTIG = Offensive_Targeted_Insult_Group, OTII - Offensive_Targeted_Insult_Individual, OT Offensive_Targeted_Insult_Other, OU = Offensive_Untargeted, NK = not-Kannada.

Table 3 gives the class-wise performance of DistilmBERT for the Kannada dataset. Our proposed method achieves the best result on 3 out of the 6 classes including the one with the most and least support (number of examples). For the "Offensive Targeted Insult Other" category, which has the least number of examples, our proposed method achieves an F1 score of 28.5% whereas the second-best (CMI + Pseudo Labeling) is 20.0%.

We also see that when every other aspect is kept the same, CMI loss performs better than focal loss in almost all cases. This shows that the level of code-mixing plays an important role in the difficulty of classification. The performance on Malayalam is very high and significantly better than on the two languages.

This can be explained by the higher inter-annotator agreement (Krippendorff's alpha = 0.89) on the Malayalam dataset, which makes it easier for the models since there is a clearer boundary between the classes. Malayalam has only 4 classes whereas the other 2 languages have 5 classes, which further makes it easier to classify Malayalam than Tamil and Kannada. For Tamil and Kannada, the inter-annotator agreement is low (Krippendorff's alpha = 0.66, 0.78 respectively), which means that the instances in these datasets have high subjectivity, confusion and overlapping boundaries between the classes. Hence, even the models find it more difficult to learn from the labelled data. The models perform better on Tamil than Kannada because the Tamil dataset is approximately six times larger than the Kannada dataset.

We conduct the Stuart-Maxwell's test [47] to check the statistical significance of our results. We compare our proposed solution (DistilmBERT + CMI loss + Cosine Normalization + Pseudo-labelling) with the baseline ensemble model from [48]. We achieve p-values 0.0005 for Kannada, 0.0003 for Tamil and 0.0376 for Malayalam. This shows that our results are statistically significant against an threshold of 0.05.

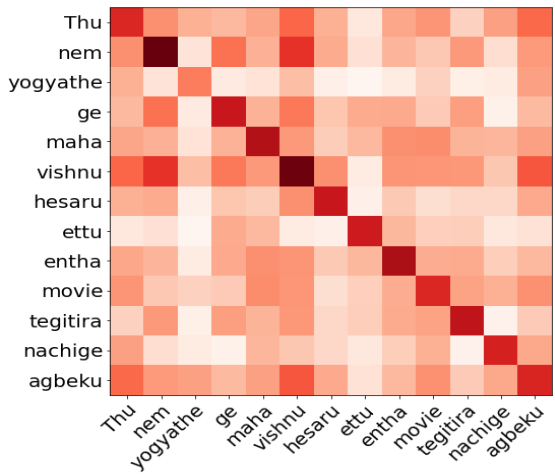

*Fig 1* Attention weights plot for a mis-classified Kannada example. Notice that the word "vishnu" is given high importance

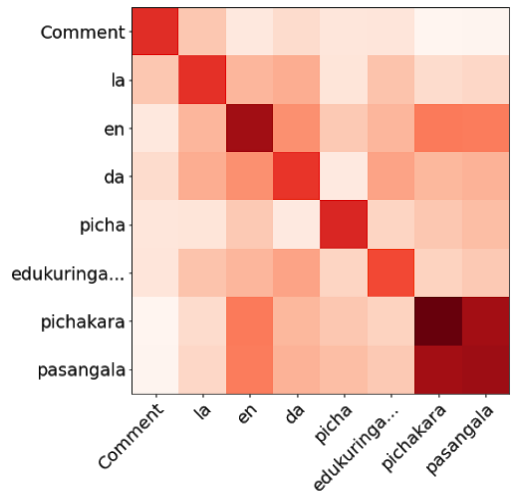

*Fig 3* Attention weights plot for an offensive example. The words "pichakara pasangala" (beggar people) are given high importance.

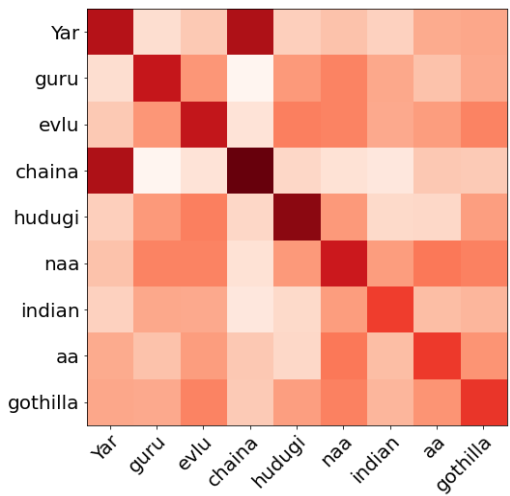

*Fig 2* Attention weights plot for a racist example. The word "chaina" (China) is given high importance.

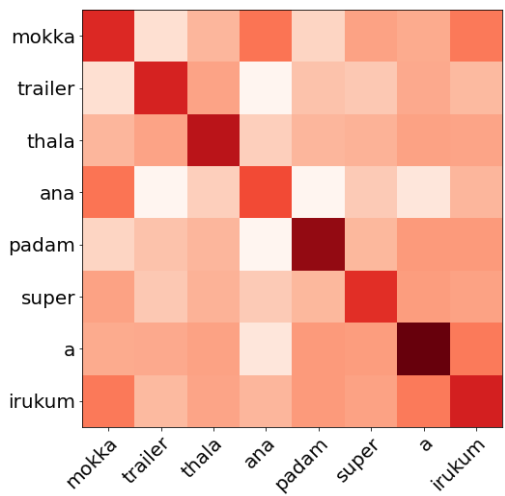

*Fig 4* Attention weights plot for a mis-classified Tamil. the words "thala", "a", "padam" are given high importance.

## 7. Error Analysis

We take the IndicBERT model and analyze some of the instances where the model did and did not predict the expected output. Table 4 gives such instances.

Figure 1 shows the attention weights plot for sentence 1 from table 4. The word "vishnu" is given high importance. The reason for this (mis)prediction could be the word "vishnu" (or other nouns) being present in a lot of sentences belonging to the "Offensive Targeted Insult Individual" class.

For sentences 2 and 6, the ground truth labels seem unlikely and the predicted labels are better. This shows that the model is robust to errors in human labelling to an extent. The unlikely ground truth labels can be attributed to the low inter-annotator agreement for Tamiland Kannada.

Sentence 3 has subtle racism. Our model correctly predicts the label "Offensive Targeted Insult Individual". Further, from its attention plot (figure 2), we can observe that the most importance is given to "chaina" (China) which is the word related to racism.

Sentence 4 in the table is offensive, but it's not clear who it is offending. The model correctly predicts "Offensive Targeted Insult Other". From Figure 3, we see that the model does a good job in figuring out that "pichakara pasangala" (beggar people) is the part it should pay attention to.

| Sl. No | Language | Input | Translated Input | Actual Label | Predicted Label |
|---|---|---|---|---|---|
| 1 | Kannada | Thu nem yogyathe ge maha vishnu hesaru ettuentha movie tegitira nachige agbeku nemge | Shame on you, you have made such a (bad) movie and named it Maha Vishnu | Offensive Untargeted | Offensive Targeted Insult Individual |
| 2 | Kannada | ಇಬ್ಬರು ಕಳ್ಳರೆ (Ibbaru Kallare) | Both of you are thieves | Offensive Targeted Insult Group | Not Offensive |
| 3 | Kannada | Yar Guru evlu chaina hudugi naaindian aa gothilla | who is this girl guru, I don't know if she's Indian or Chinese | Offensive Targeted Insult Individual | Offensive Targeted Insult Individual |
| 4 | Tamil | Comment la en da picha edukuringa... pichakara pasangala | Why are they begging in the comment section… beggars | Offensive Targeted Insult Group | Offensive Targeted Insult Other |
| 5 | Tamil | வேற லெவல் சூப்பர் ஹீரோ படம் போல வாழ்த்துக்கள் ரவி கோமாளி (Vera level super hero padam pole vaazhthukkal ravi comali) | the movie was on a diff level, congrats ravi comali (joker) | Not Offensive | Offensive Targeted Insult Individual |
| 6 | Tamil | For Haters அசுரன் பரியேறும் பெருமாள் மெட்ராஸ் அட்டகத்தி காலா இந்த படம் எடுக்கும் போது எங்க போயிருந்தீங்க..... (For Haters Asuran, pari erum perumal, madras, attakaththi, kala intha padam edukum pothu enga poirundeenge) | For haters, movies asuran, pari erum perumal, madras, kala, when he was directing/taking all these where were you? | Not Offensive | Offensive Untargeted |
| 7 | Tamil | mokka trailer thala ana padam super airukum | boring trailer thala but the movie will be superb | Not Offensive | Offensive Targeted Insult Individual |

Table 4 Predicted label and actual label for some examples from Kannada and Tamil datasets.

Sentence 5 is not offensive but it is predicted as 'offensive targeted insult individual'. This is because the word "comali" here refers to a movie but in a literal sense means 'joker' which could be used to offend someone.

Sentence 7 is a not offensive instance predicted as "Offensive targeted insult individual". From its attention weights plot (Figure 4), we see that high importance was given to the words "a", "padam" and "thala". A possible reason for the misprediction could be that the word "thala" appeared in many "Offensive targeted insult individual" sentences (since the word is used to describe a movie actor).

## 8. Conclusion and Future Work

In this paper, we introduce a novel CMI based focal loss that leverages the code-mixed phrases present in the data to better predict offensive language in Dravidian code-mixed sentences. We also notice that augmenting this with cosine normalization helps improve the model. To better understand our results, we compare them both quantitatively and qualitatively. Our method is language-agnostic and can therefore be extended toother low resource languages. Our additions (CMI loss and Cosine Normalization) are simple yet effective and could be augmented to other models and approaches with minimal effort. Additionally, constructing synthetic code-mixed data and the usage of language-specific tokenizers for the multi-lingual models are potential future research directions.

## 9. Declarations

No funding and no conflict of interest.

## 10. References

1. Akiwowo, S., Vidgen, B., Prabhakaran, V., Waseem, Z. (eds.): Proceedings of the Fourth Workshop on Online Abuse and Harms. Association for Computational Linguistics (2020). URL https://www.aclweb.org/anthology/2020.alw-1.0
2. Bali, K., Sharma, J., Choudhury, M., Vyas, Y.: "i am borrowing ya mixing?" an analysis of english-hindi code mixing in facebook. In: Proceedings of the First Workshop on Computational Approaches to Code Switching (2014)
3. Barman, U., Das, A., Wagner, J., Foster, J.: Code mixing: A challenge for language identification in the language of social media. In: Proceedings of the First Workshop on Computational Approaches to Code Switching. Association for Computational Linguistics (2014). URL https://www.aclweb.org/anthology/W14-3902
4. Basile, V., Bosco, C., Fersini, E., Nozza, D., Patti, V., Rangel Pardo, F.M., Rosso, P., Sanguinetti, M.: SemEval-2019 task 5: Multilingual detection of hate speech against immigrants and women in Twitter. In: Proceedings of the 13th International Workshop on Semantic Evaluation, pp. 54–63. Association for Computational Linguistics, Minneapolis, Min-Minnesota, USA (2019). DOI 10.18653/v1/S19-2007. URL https://aclanthology.org/S19-2007
5. Chakravarthi, B.R., Jose, N., Suryawanshi, S., Sherly, E., McCrae, J.P.: A sentiment analysis dataset for code-mixed Malayalam-English. In: Proceedings of the 1st Joint Workshop on Spoken Language Technologies for Under-resourced languages (SLTU) and Collaboration and Computing for Under-Resourced Languages (CCURL). European Language Resources Association (2020). URL https://www.aclweb.org/anthology/2020.sltu-1.25
6. Chakravarthi, B.R., Muralidaran, V., Priyadharshini, R., McCrae, J.P.: Corpus creation for sentiment analysis in code-mixed Tamil-English text. In: Proceedings of the 1st Joint Workshop on Spoken Language Technologies for Under-resourced languages (SLTU) and Collaboration and Computing for Under-Resourced Languages (CCURL). European Language Resources Association (2020). URL https://www.aclweb.org/anthology/202 0.sltu-1.28
7. Chakravarthi, B.R., Priyadharshini, R., Jose, N., M, A.K., Mandl, T., Kumaresan, P.K., Ponnusamy, R., V, H., Sherly, E., McCrae, J.P.: Findings of the shared task on Offensive Language Identification in Tamil, Malayalam, and Kannada. In: Proceedings of the First Workshop on Speech and Language Technologies for Dravidian Languages. Association for Computational Linguistics (2021)
8. Chen, Y., Skiena, S.: False-friend detection and entity matching via unsupervised transliteration. arXiv preprint arXiv:1611.06722 (2016).
9. Chittaranjan, G., Vyas, Y., Bali, K., Choudhury, M.: Word-level language identification using CRF: Code-switching shared task report of MSR India system. In: Proceedings of the First Workshop on Computational Approaches to Code Switching, pp. 73–79. Association for Computational Linguistics (2014). DOI 10.3115/v1/W14-3908. URL https://www.aclweb.org/anthology/W14-3908
10. Corazza, M., Menini, S., Cabrio, E., Tonelli, S., Villata, S.: A multilingual evaluation for online hate speech detection. ACM Trans. Internet Technol. **20**(2) (2020). DOI 10.1145/3377323. URL https://doi.org/10.1145/3377323
11. Das, A., Gambäck, B.: Identifying languages at the word level in code-mixed Indian social media text. In: Proceedings of the 11th International Conference on Natural Language Processing. NLP Association of India, Goa, India (2014). URL https://aclanthology.org/W14-5152
12. Devlin, J., Chang, M.W., Lee, K., Toutanova, K.: Bert: Pre-training of deep bidirectional transformers for language understanding. arXiv preprint arXiv:1810.04805 (2018)
13. Dowlagar, S., Mamidi, R.: Cmsaone@dravidian-codemix-fire2020: A meta embedding and transformer model for code-mixed sentiment analysis on social media text (2021)
14. Fortuna, P., Nunes, S.: A survey on automatic detection of hate speech in text. ACM Comput. Surv. **51**(4) (2018). DOI 10.1145/3232676. URL https://doi.org/10.1145/3232676
15. Gambäck, B., Das, A.: On measuring the complexity of code-mixing. In: Proceedings of the 11th International Conference on Natural Language Processing (2014)
16. Gidaris, S., Komodakis, N.: Dynamic few-shot visual learning without forgetting (2018)
17. Hande, A., Priyadharshini, R., Chakravarthi, B.R.: KanCMD: Kannada CodeMixed dataset for sentiment analysis and offensive language detection. In: Proceedings of the Third Workshop on Computational Modeling of People's Opinions, Personality, and Emotion's in Social Media. Association for Computational Linguistics (2020). URL https://www.aclweb.org/anthology/2020.peoples-1.6
18. Kakwani, D., Kunchukuttan, A., Golla, S., N.C., G., Bhattacharyya, A., Khapra, M.M., Kumar, P.: Indicnlpsuite: Monolingual corpora, evaluation benchmarks and pre-trained multilingual language models for Indian languages. In: Findings of the Association for Computational Linguistics: EMNLP 2020, pp. 4948–4961. Association for Computational Linguistics (2020). URL https://aclanthology.org/2020.findings-emnlp.445

19. Kang, B., Xie, S., Rohrbach, M., Yan, Z., Gordo, A., Feng, J., Kalantidis, Y.: Decoupling representation and classifier for long-tailed recognition (2020).

20. Kumar, R., Bhanodai, G., Pamula, R., Chennuru, M.R.: TRAC-1 shared task on aggression identification: IIT(ISM)@COLING'18. In: Proceedings of the First Workshop on Trolling, Aggression and Cyberbullying (TRAC-2018). Association for Computational Linguistics (2018). URL https://www.aclweb.org/anthology/W18-4407

21. Kumar, R., Ojha, A.K., Malmasi, S., Zampieri, M.: Benchmarking aggression identification in social media. In: Proceedings of the First Workshop on Trolling, Aggression and Cyberbullying (TRAC-2018), pp. 1–11. Association for Computational Linguistics, Santa Fe, New Mexico, USA (2018). URL https://aclanthology.org/W18-4401

22. Kumar, R., Ojha, A.K., Malmasi, S., Zampieri, M.: Evaluating aggression identification in social media. In: Proceedings of the Second Workshop on Trolling, Aggression and Cyberbullying. European Language Resources Association (ELRA) (2020). URL https://www.aclweb.org/anthology/2020.trac-1.1

23. Laub, Z.: Hate speech on social media: Global comparisons. Council on Foreign Relations (2019). URL https://www.cfr.org/backgrounder/hate-speech-social-media-global-comparisons

24. Lin, T.Y., Goyal, P., Girshick, R., He, K., Dollár, P.: Focal loss for dense object detection. In: Proceedings of the IEEE international conference on computer vision, pp. 2980–2988 (2017)

25. Liu, R., Xu, G., Vosoughi, S.: Enhanced offensive language detection through data augmentation (2020)

26. Liu, Z., Miao, Z., Zhan, X., Wang, J., Gong, B., Yu, S.X.: Large-scale long-tailed recognition in an open world (2019)

27. Ma, Y., Zhao, L., Hao, J.: XLP at SemEval-2020 Task 9: Cross-lingual models with the focal loss for sentiment analysis of code-mixing language. In: Proceedings of the Fourteenth Workshop on Semantic Evaluation, pp. 975–980. International Committee for Computational Linguistics (2020).

28. Mahata, S.K., Das, D., Bandyopadhyay, S.: Junlp@dravidian-codemix-fire2020: Sentiment classification of code-mixed tweets using bi-directional rnn and language tags (2020).

29. Manolescu, M., Löfflad, D., Saber, A.N.M., Tari, M.M.: Tueval at semeval-2019 task 5: LSTM approach to hate speech detection in English and Spanish. In: J. May, E. Shutova, A. Herbelot, X. Zhu, M. Apidianaki, S.M. Mohammad (eds.) Proceedings of the 13th International Workshop on Semantic Evaluation, SemEval@NAACL-HLT 2019, Minneapolis, MN, USA, June 6-7, 2019, pp. 498–502. Association for Computational Linguistics (2019). DOI 10.18653/v1/s19-2089. URL https://doi.org/10.18653/v1/s19-2089

30. Mathur, P., Shah, R., Sawhney, R., Mahata, D.: Detecting offensive tweets in Hindi-English code-switched language. In: Proceedings of the Sixth International Workshop on Natural Language Processing for Social Media, pp. 18–26. Association for Computational Linguistics, Melbourne, Australia (2018). DOI 10.18653/v1/W18-3504. URL https://aclanthology.org/W18-3504

31. Patra, B.G., Das, D., Das, A., Prasath, R.: Shared task on sentiment analysis in indian languages (sail) tweets - an overview. In: R. Prasath, A.K. Vuppala, T. Kathirvalavakumar (eds.) Mining Intelligence and Knowledge Exploration. Springer International Publishing (2015)

32. Patwa, P., Aguilar, G., Kar, S., Pandey, S., PYKL, S., Gambäck, B., Chakraborty, T., Solorio, T., Das, A.: Semeval-2020 task 9: Overview of sentiment analysis of code-mixed tweets. In: Proceedings of the 14th International Workshop on Semantic Evaluation (SemEval-2020). Association for Computational Linguistics (2020)

33. Patwa, P., Bhardwaj, M., Guptha, V., Kumari, G., Sharma, S., PYKL, S., Das, A., Ekbal, A., Akhtar, S., Chakraborty, T.: Overview of constraint 2021 shared tasks: Detecting English covid-19 fake news and Hindi hostile posts. In: Proceedings of the FirstWorkshop on Combating Online Hostile Posts in Regional Languages during Emergency Situation (CONSTRAINT). Springer (2021)

34. Patwa, P., Pykl, S., Das, A., Mukherjee, P., Pulabaigari, V.: Hater-O-genius aggression classification using capsule networks. In: Proceedings of the 17th International Conference on Natural Language Processing (ICON), pp. 149–154. NLP Association of India (NLPAI), Indian Institute of Technology Patna, Patna, India (2020). URL https://aclanthology.org/2020.icon-main.19

35. Pratapa, A., Bhat, G., Choudhury, M., Sitaram, S., Dandapat, S., Bali, K.: Language modeling for code-mixing: The role of linguistic theory based synthetic data. In: Proceedings of the 56th Annual Meeting of the Association for Computational Linguistics (Volume 1: Long Papers). Association for Computational Linguistics, Melbourne, Australia (2018). DOI 10.18653/v1/P18-1143. URL https://aclanthology.org/P18-1143

36. Qi, H., Brown, M., Lowe, D.G.: Low shot learning

with imprinted weights (2018)
37. Raha, T., Roy, S.G., Narayan, U., Abid, Z., Varma, V.: Task adaptive pretraining of transformers for hostility detection (2021)
38. Ranasinghe, T., Zampieri, M.: An evaluation of multilingual offensive language identification methods for the languages of india. Information 12(8) (2021)
39. Risch, J., Krestel, R.: Bagging BERT models for robust aggression identification. In: Proceedings of the Second Workshop on Trolling, Aggression and Cyberbullying. European Language Resources Association (ELRA) (2020). URL https://www.aclweb.org/anthology/2020.trac-1.9
40. Risch, J., Stoll, A., Ziegele, M., Krestel, R.: hpidedis at germeval 2019: Offensive language identification using a german BERT model. In: Proceedings of the 15th Conference on Natural Language Processing, KONVENS 2019, Erlangen, Germany, October 9-11, 2019 (2019). URL https://corpora.linguistik.uni-erlangen.de/data/konvens/proceedings/papers/germeval/Germeval\_Task\_2\_2019\_paper\_10.HPIDEDIS.pdf
41. Sabour, S., Frosst, N., Hinton, G.E.: Dynamic routing between capsules. In: Proceedings of the 31st International Conference on Neural Information Processing Systems, NIPS'17. Curran Associates Inc., Red Hook, NY, USA (2017)
42. Safi Samghabadi, N., Patwa, P., PYKL, S., Mukherjee, P., Das, A., Solorio, T.: Aggression and misogyny detection using BERT: A multi-task approach. In: Proceedings of the Second Workshop on Trolling, Aggression and Cyberbullying. European Language Resources Association (ELRA) (2020). URL https://aclanthology.org/2020.trac-1.20
43. Sai, S., Sharma, Y.: Towards offensive language identification for Dravidian languages. In: Proceedings of the First Workshop on Speech and Language Technologies for Dravidian Languages. Association for Computational Linguistics, Kyiv (2021). URL https://aclanthology.org/2021.dravidianlangtech-1.3
44. Sanh, V., Debut, L., Chaumond, J., Wolf, T.: Distilbert, a distilled version of bert: smaller, faster, cheaper and lighter. arXiv preprint arXiv:1910.01108 (2019)
45. Stammbach, D.: Offensive language detection with neural networks for germeval task 2018 (2018)
46. Steimel, K., Dakota, D., Chen, Y., Kübler, S.: Investigating multilingual abusive language detection: A cautionary tale. In: Proceedings of the International Conference on Recent Advances in Natural Language Processing (RANLP 2019). INCOMA Ltd. (2019). URL https://aclanthology.org/R19-1132
47. STUART, A.: A TEST FOR HOMOGENEITY OF THE MARGINAL DISTRIBUTIONS IN A TWO-WAY CLASSIFICATION. Biometrika 42(3-4) (1955). DOI 10.1093/biomet/42.3-4.412. URL https://doi.org/10.1093/biomet/42.3-4.412
48. Tula, D., Potluri, P., Ms, S., Doddapaneni,S., Sahu, P., Sukumaran, R., Patwa, P.: Bitions@DravidianLangTech-EACL2021: Ensemble of multilingual language models with pseudo labelling for offence detection in Dravidian languages. In: Proceedings of the First Workshop on Speech and Language Technologies for Dravidian Languages, pp. 291–299. Association for Computational Linguistics, Kyiv (2021). URL https://aclanthology.org/2021.dravidianlangtech-1.42
49. Vaswani, A., Shazeer, N., Parmar, N., Uszkoreit, J., Jones, L., Gomez, A.N., Kaiser, L., Polosukhin, I.: Attention is all you need. In: I. Guyon, U. von Luxburg, S. Bengio, H.M. Wallach, R. Fergus, S.V.N. Vishwanathan, R. Garnett (eds.) Advancesin Neural Information Processing Systems 30: Annual Conference on Neural Information Processing Systems 2017, December 4-9, 2017, Long Beach, CA, USA, pp. 5998–6008 (2017). URL https://proceedings.neurips.cc/paper/2017/hash/3f5ee243547dee91fbd053c1c4a845aa-Abstract.html
50. Vyas, Y., Gella, S., Sharma, J., Bali, K., Choudhury, M.: POS tagging of English-Hindi code-mixed social media content. In: Proceedings of the 2014 Conference on Empirical Methods in Natural Language Processing (EMNLP). Association for Computational Linguistics (2014). DOI 10.3115/v1/D14-1105. URL https://aclanthology.org/D14-1105
51. Waseem, Z., Chung, W.H.K., Hovy, D., Tetreault, J. (eds.): Proceedings of the First Workshop on Abusive Language Online. Association for Computational Linguistics (2017). URL https://www.aclweb.org/anthology/W17-3000
52. Zampieri, M., Malmasi, S., Nakov, P., Rosenthal, S., Farra, N., Kumar, R.: Semeval-2019 task 6: Identifying and categorizing offensive language in social media (offenseval) (2019)
53. Zampieri, M., Nakov, P., Rosenthal, S., Atanasova, P., Karad (Barman, 2014)zhov, G., Mubarak, H., Derczynski, L., Pitenis, Z., Çöltekin, Ç.: SemEval-2020 task 12: Multilingual offensive language identification in social media (OffensEval 2020). In: Proceedings of the Fourteenth Workshop on Semantic Evaluation. International Committee for Computational Linguistics (2020).